\documentclass[11pt,a4paper]{article}
\usepackage[hyperref]{acl2019}
\usepackage{times}
\usepackage{latexsym}
\usepackage{amsmath}
\usepackage{amssymb}
\usepackage{diagbox}
\usepackage{enumitem}
\usepackage{graphicx}
\usepackage{multirow}
\usepackage{tabularx}
\usepackage{booktabs}
\usepackage{url}
\usepackage{amsmath}

\aclfinalcopy 

\newcommand{\cev}[1]{\reflectbox{\ensuremath{\vec{\reflectbox{\ensuremath{#1}}}}}}
\title{Improving Neural Question Generation using World Knowledge}

\author{Deepak Gupta$^{\ast}$, Kaheer Suleman$^{\dagger}$, Mahmoud Adada$^{\dagger}$, Andrew McNamara$^{\dagger}$ and Justin Harris$^{\dagger}$ \\
  $^{\ast}$Indian Institute of Technology Patna, India  \\
 $^{\dagger}$Microsoft Research Montreal, Canada \\
  {\tt 
  $^{\ast}$deepak.pcs16@iitp.ac.in}\\
  {\tt $^{\dagger}$\{kasulema, Mahmoud.Adada, Andrew.McNamara, Justin.Harris\} @{microsoft.com} 
  }
  }

\date{}

\begin{document}
\maketitle
\begin{abstract}
In this paper, we propose a method for incorporating world knowledge (linked entities and fine-grained entity types) into a neural question generation model. This world knowledge helps to encode additional information related to the entities present in the passage required to generate human-like questions. We evaluate our models on both SQuAD and MS MARCO to demonstrate the usefulness of the world knowledge features. The proposed world knowledge-enriched question generation model is able to outperform the vanilla neural question generation model by $1.37$ and $1.59$ absolute BLEU\_4 score on SQuAD and MS MARCO test datasets, respectively. 
\end{abstract}
\section{Introduction}
The task of question generation (QG) aims to generate syntactically and semantically sound questions from a given text for which a given answer would be a correct response. Recently, there has been increased research interest in the QG task due to \textbf{(1)} the wide success of neural network based sequence-to-sequence techniques \cite{sutskever2014sequence} for various NLP tasks and \cite{bahdanau2014neural,srivastava2015unsupervised,xu2015show, rush2015neural,kumar2016ask}, \textbf{(2)} the abundance of large question answering datasets: SQuAD \cite{squad}, NewsQA \cite{trischler2016newsqa}, MS MARCO \cite{ms-marco}). \begin{table}[h]
\centering
\begin{tabularx}{\linewidth}{X}
\textbf{Text:} Kevin Hart and Dwayne Johnson are taking on MTV. The duo, who co-star in the upcoming film Central Intelligence, are set to co-host the 2016 MTV Movie Awards, The Hollywood Reporter has confirmed.
\\ 
\textbf{Ans:} \textit{Central Intelligence}\\
\textbf{Q$_{gen}$:} Dwayne Johnson co-starred with Kevin Hart in what \textcolor{red}{organization}? \\
\textbf{Q$_{hum}$}: In which  \textcolor{blue}{film} did Dwayne Johnson collaborate with comedian Kevin Hart? \\
\bottomrule 
\end{tabularx}
\caption{Sample texts, along with the machine (\textbf{Q$_{gen}$}) and human (\textbf{Q$_{hum}$})  generated questions for the given answer (\textbf{Ans}). In the machine-generated questions, where the corresponding entities could not be resolved, are shown in \textcolor{red}{red}, the corresponding resolved entities in human-generated questions are in \textcolor{blue}{blue}.}
\label{table:intro_example}
\end{table}

In this paper, we advocate for improving question generation systems using world knowledge, which has not been investigated as of yet. We explore world knowledge in the form of entities present in text and exploit the associated entity knowledge to generate human-like questions. In our experiments, we use two types of world knowledge: linked entities to the Wikipedia knowledge base and fine-grained entity types (FGET). Table \ref{table:intro_example} illustrates how this form of world knowledge can be used to improve question generation. Here, ``\textit{Central Intelligence}'' is the name of a movie, and `\textit{Dwayne Johnson}' and `\textit{Kevin Hart}' are actors. The world knowledge here is the name of a movie (``\textit{Central Intelligence}''), which helps the model to generate the correct word `\textit{film}' instead of the incorrect word `\textit{organization}'.

\indent We adopt the sequence-to-sequence model \cite{bahdanau2014neural} equipped with the copy mechanism \cite{pointing-unknown-gulcehre-2016,go-to-the-point-see-2017} as our base model for question generation. The entity linking and fine-grained entity typing information are fed to the network along with the answer of interest. We believe this is the first work that explores world-knowledge in the form of linked entities and fine-grained entity types as features to improve neural question generation models. 

\section{Related Work} 
Previous work on question generation can be categorized into two types : rule-based methods and neural network-based methods. The work in this paper belongs to the latter category. Traditional rule-based approaches involve the manual formulation of templates to generate questions and further refinement of the generated questions using features based on semantic information \cite{lindberg2013generating, heilman2010good}, ontologies \cite{labutov2015deep}, argument structures \cite{chali2015towards}, etc. 

\indent Recently, work on question generation has drifted towards neural-based approaches. These approaches typically involve end-to-end supervised learning to generate questions. \citet{learning-to-ask} proposed sequence-to-sequence learning for question generation from text passages. \citet{zhou2017neural} utilized the answer-position, and linguistic features such as named entity recognition (NER) and parts of speech (POS) information to further improve the QG performance as the model is aware of which answer a question needs to be generated. In the work of  \citet{wang2016multi} a multi-perspective context matching algorithm is employed. 
\citet{harrison2018neural} use a set of rich linguistic features along with an NQG model. \cite{context-multi-perspective-qg} used the matching algorithm proposed by \cite{wang2016multi} to compute the similarity between the
target answer and the passage for collecting relevant contextual information under the different perspectives, so that contextual
information can be better considered by the encoder.
More recently, \citet{kim2018improving} has claimed to improve the performance of the QG model by replacing the target answer in the original passage with special tokens. Other NQG models include \cite{para-level-max-out-qg,ans-focused-qg,gao2018difficulty}, which generate questions mainly from the SQuAD and MS MARCO datasets. There is another line of research \cite{bahri2024consumer,yadav2021nlm,yadav2021reinforcement,yadav2022chq,yadav2022question,yadav2022towards,yadav2023towards} in which the question is generated as the summary of the long-form question with the goal of providing an accurate answer \cite{zhao2024heterogeneous,chaturvedi2024aspect,naik2024no,agarwal2025overview,bhattacharya2022lchqa}. 

\section{Proposed Approach}

\subsection{World Knowledge Enrich Encoder}
Our proposed model is based on the sequence-to-sequence \cite{bahdanau2014neural} paradigm. For the encoder, we utilize a Long Short Term Memory (LSTM) \cite{hochreiter1997long} network. In order to capture more contextual information, we use a two-layer bidirectional LSTM (Bi-LSTM). Inspired by the success of using linguistic features in \cite{zhou2017neural,harrison2018neural}, we exploit word knowledge in the form of entity linking and fine-grained entity typing in the encoder of the network. The Bi-LSTM encoder reads the passage words and their associated world knowledge features (c.f. section \ref{entity-linking}, \ref{fgec}) to produce a sequence of word-and-feature vectors. We concatenate the word vectors, the embedded world knowledge feature vectors and the answer position indicator embedding vector as the input to the Bi-LSTM encoder. 
\subsubsection{Entity Linking}
\label{entity-linking}
In previous works \cite{zhou2017neural,harrison2018neural}, named entity type features have been used. These features, however, only allow for the encoding of coarse-level information, such as knowledge of if an entity belongs to a set of predefined categories, such as `\textit{PERSON}', `\textit{LOCATION}', and `\textit{ORGANIZATION}'. To alleviate this, we use the knowledge in the form of linked entities. In our experiments, we use Wikipedia as the knowledge base for which to link entities.  This specific task (also known as Wikification \cite{cheng2013relational}) is the task of identifying concepts and entities in text and disambiguation them into the most specific corresponding Wikipedia pages. We followed the approach by \citet{cheng2013relational} for the Wikification.
\indent 
The Wikification process is performed on the input passage $P$ having $n$ words $\{w_1^p, w_2^p, \ldots, w_n^p \}$, we map each word of the passage to their corresponding Wikipedia title to generate a sequence of Wikipedia titles $E_{link}=\{e_1^{p}, e_2^p, \ldots, e_n^p \}$. For multi-word mentions, we assign the same Wikipedia title to each word of the mention. In order to project the word and entity in the same vector space, we jointly learn pre-trained word-entity vector embeddings using the method proposed by \citet{yamada2016joint}.  
\subsubsection{Fine-grained Entity Types (FGET)} \label{fgec}
Fine-grained entity typing consists of assigning types from a hierarchy to entity mentions in text. Similar to the approach in \cite{fgec}, we build a classification model to classify the predicted entity mentions from the entity linker, discussed in section \ref{entity-linking}, into one of the predefined fine-grained entity types ($112$ entities) \cite{ling2012fine}. 
The inputs to the network are a sub-sequence of passage sentence $S=(w_1, w_2, \dots, w_T)$ and the target entity $M = (w_{p}, \dots, w_{t})$ ($p, t \in [1,T], p \le t$) of length $t-p+1$. The sub-sequence $S$ is a context sentence of length $T$ for the given mention $M$, where $M \in S$.  Using the FGET classification approach discussed in \cite{fgec}, we obtain the representation $R$ of the passage sentence $S$.
Thereafter, a soft-max layer is employed to obtain the probability distribution over the set of fine-grained entity types $Y$. Concretely,
\begin{gather}
\label{eq2}
 prob(y|M, S) = \text{softmax}(\mathbf{W}R + b) \\
\hat y = \arg \max_{y\in Y}  prob (y|M, S)
\end{gather}
where weight matrix $\mathbf{W}$ is treated as the \textbf{learned type embedding} and $b$ is the bias.



Similar to the process we use for the linked entities, we map the passage words to their corresponding fine-grained entity types to get a sequence of FGET $E_{fget}=\{f_1^{p}, f_2^p, \ldots, f_n^p \}$. The final embedding of a word at a given time $t$ of the passage $P$, is computed as:
\begin{equation}
    x_t= \hat{a}_t^{p} \oplus \hat{w}_t^{p}\oplus \hat{e}_t^{p} \oplus \hat{f}_t^{p}
\end{equation}
where, $\hat{a}_t^{p}$  $\hat{w}_t^{p}$,  $\hat{e}_t^{p}$ and  $\hat{f}_t^{p}$ are the embeddings of the answer position, word, linked entity, and fine-grained entity type of the token $t$ of the passage. The final embedding sequence $x = (x_{1}, x_{2}, \dots, x_{n})$, is passed to a Bi-LSTM encoder to produce two sequences of hidden vectors, the forward sequence $ (\vec{h}_{1}, \vec{h}_{2}, \dots, \vec{h}_{n}) $ and the backward sequence $ (\cev{h}_{1}, \cev{h}_{2}, \dots, \cev{h}_{n}) $.
Lastly, the output sequence of the encoder is the concatenation of the two sequences, $ h_{i} = [\vec{h}_{i} \oplus \cev{h}_{i}] $.

\subsection{Decoding with Attention}
We use a two-layer LSTM for the decoder. Words are generated sequentially conditioned on the encoder output and the previous decoder step. Formally, at decoding time step $t$, the LSTM decoder reads the previous word embedding $ y_{t-1} $ and context vector $ c_{t -1} $  to compute the new hidden state $ d_{t} $. The context vector $ c_{t} $ at time step $t$ is computed using the attention mechanism in \cite{luong-attention}, which matches the current decoder state $d_t$ with each encoder hidden state $h_i$ to get a relevance score. 
A one-layer feed-forward network takes the decoder state $s_t$ and $c_t$ and predicts the probability distribution over the decoder vocabulary. Similar to \cite{ans-focused-qg,context-multi-perspective-qg}, we also use the copy mechanism from \cite{pointing-unknown-gulcehre-2016} to deal with the rare and unknown words.

\section{Experimental Results}
We evaluated the performance of our approach on SQuAD \cite{squad} and MS MARCO v2.1 \cite{ms-marco}. SQuAD is composed of more than 100K questions 
posed by crowd workers on $536$ Wikipedia articles. We used the same split as \cite{zhou2017neural}. 
\indent MS MARCO dataset contains $1$ million queries with corresponding answers and passages. All questions are sampled from real anonymized user queries, and context passages are extracted from real web documents. We picked a subset of MS
MARCO data where answers ($<=10$ words) are sub-spans within the passages ($<=600$ words), and use the dev set as the test set ($7,849$), and split the train set with a ratio of 90\%-10\% into train ($1,36,337$) and dev ($15,148$) sets.
\subsection{Experimental Settings}
We optimized network hyperparameters for both datasets via their respective development set. The LSTM cell hidden size was 512 for both datasets.  We used a $500$-dimension vector\footnote{\url{https://bit.ly/2TqrJh4}} jointly trained for word and Wikipedia entity from \cite{yamada2016joint} for the pretrained word and entity embeddings. The dimensions of answer tagging and entity type tagging were set to $100$. The model was optimized via gradient descent using the Adam \cite{kingma2014adam} optimiser with a learning rate of $0.001$ and mini-batch size 64.  We selected the models with the highest BLEU-4 \cite{papineni2002bleu} scores as our final models. At inference time, we used beam search with a beam size of $12$ (also optimized on the dev set) for the SQuAD dataset, and greedy search was adopted for the MS MARCO dataset, as it performed near the best result compared to beam search. For both datasets, we restrict the target vocabulary to the most frequent $20,000$ words. We evaluate the question generation performance in terms of BLEU \cite{papineni2002bleu}, METEOR \cite{meteor}, and ROUGE-L \cite{lin2004rouge}, and using the evaluation package released by \citet{sharma2017nlgeval}.
\section{Results and Analysis}
\begin{table*}[h]
\resizebox{\linewidth}{!}{%
\begin{tabular}{l|l|l|l|l|l|l}
\hline
\multirow{2}{*}{\textbf{\diagbox{Model}{Dataset}}} & \multicolumn{3}{c|}{\textbf{SQuAD}} & \multicolumn{3}{c}{\textbf{MS MARCO}} \\ \cline{2-7} 
 & BLEU\_4 & METEOR & ROUGE-L & BLEU\_4 & METEOR & ROUGE-L \\ \hline
s2s+Att & 7.53 & 13.38 & 33.98 & 8.86 &13.98  & 34.57 \\ 
NQG & 12.54 & 17.67 & 41.74 & 11.73 & 18.06 & 37.64 \\ \hline
NQG+EL & 11.78 & 17.41 & 40.02 &  11.52&	18.34&	37.56  \\ 
NQG+ EL (pre) & 13.28 & 18.03 & 41.89 & 12.95 & 20.07 & 39.76 \\ 
NGQ+FGET & \textbf{13.91} & 18.51 & \textbf{42.53} & 12.01 & 18.82 & 37.88 \\ 
NGQ+FGET (pre) & \textbf{13.91} & 18.48 & 42.46 &12.95 &	20.07	&39.76  \\ 
NQG+ EL (best) + FGET (best) & 13.69 & 18.50 & 42.13 & \textbf{13.32} & \textbf{20.47} & \textbf{40.05} \\ \hline \hline
NQG + NER  & 13.22 & 18.54 & 41.36  & 12.18 & 18.46 & 38.04 \\ 
NQG + NER + FGET  &  12.73 & 18.51 & 40.39 & 12.11 & 18.52 & 37.69 \\ 
NGQ + NER + FGET (pre)   &13.44  &\textbf{19.14}  & 41.27 & 12.00 & 18.56 & 37.70 \\ \hline \hline
\cite{zhou2017neural} & $13.29$ &-  &-  & - & - & - \\ \hline
 \cite{context-multi-perspective-qg} &$\textbf{13.91}$  & -  & -   &- &-& -\\ \hline
\end{tabular}%
}
\caption{Performance comparison of the proposed model on the test set of both datasets. The term \textit{`best'} refers to the best performance on the development set.}
\label{table:test-results}
\end{table*}

\begin{table*}[h]
\resizebox{\textwidth}{!}{%
\begin{tabular}{l|l|l|l|l|l|l|l|l|l|l|l|l}
\hline
\multirow{2}{*}{\diagbox{Model}{Dataset}} & \multicolumn{6}{c|}{\textbf{SQuAD}} & \multicolumn{6}{c}{\textbf{MS MARCO}} \\ \cline{2-13} 
 & BLEU\_1 & BLEU\_2 & BLEU\_3 & BLEU\_4 & METEOR & ROUGE-L & BLEU\_1 & BLEU\_2 & BLEU\_3 & BLEU\_4 & METEOR & ROUGE-L \\ \hline
s2s + Att & 30.16 & 16.80 & 11.10 & 07.77 & 13.48 & 33.96 & 31.67 &16.25  & 11.63 & 8.41 & 14.01  & 33.54  \\ \hline
NQG & 37.72 & 23.89 & 16.93 & 12.46 & 17.78 & 42.02 & 39.41 & 26.38 & 17.81 & 12.06 & 18.82 & 38.19 \\ \hline
NQG + EL & 38.57 & 23.85 & 16.65 & 12.12 & 17.47 & 40.24 & 40.84 & 27.92 & 19.27 & 13.51 & 20.34 & 39.79 \\ \hline
NQG + EL (pre) & 38.81 & 24.89 & 17.88 & 13.40 & 18.08 & 41.90& 42.01 & 28.66 & 19.81 & 13.93 & 20.96 & 40.60 \\ \hline
NGQ + FGET & 39.64 & 25.72 & 18.64 & 14.05 & 18.61 & 42.88 & 39.97 & 27.05 & 18.47 & 12.64 & 19.59 & 38.54 \\ \hline
NGQ + FGET (pre) & 40.21 & 26.17 & 19.03 & 14.41 & 18.72 & 42.90 & 39.83 & 26.84 & 18.22 & 12.33 & 19.15 & 38.40 \\ \hline
\multicolumn{1}{l|}{NQG+ EL (pre) + FGET (pre)} & 39.82 & 25.63 & 18.41 & 13.79 & 18.57 & 42.40 & 41.92 & 28.57 & 19.72 & 13.82 & 20.84 & 40.62 \\ \hline \hline 
NQG + NER & 42.05 & 26.16 & 18.05 & 12.96 & 18.51 & 41.37 & 39.99 & 27.00 & 18.38 & 12.56 & 19.22 & 38.73 \\ \hline
NQG + NER + FGET & 42.49 & 25.85 & 17.67 & 12.59 & 18.50 & 40.34 & 39.97 & 27.05 & 18.47 & 12.60 & 19.37 & 38.65 \\ \hline
NGQ + NER + FGET (pre) & 42.78 &26.68  &18.48  &13.30  &19.15  & 41.22 & 40.05 & 27.09 & 18.46 & 12.60 & 19.57 & 38.68 \\ \hline \hline
\end{tabular}%
}
\caption{Performance comparison of the proposed model with the other baselines and the state-of-the-art model on the development set of both the datasets}
\label{table:dev-results}
\end{table*}

We conducted several experiments as follows:\\
\textbf{(1) s2s+Att}: Baseline encoder-decoder based seq2seq network with attention mechanism.\\
\textbf{(2) NQG}: Extension of \textit{s2s+Att} with answer position feature.\\
\textbf{(3) NQG + EL}: Extension of \textit{NQG} with the entity linking feature (500 dimension) discussed in Section \ref{entity-linking}.\\
\textbf{(4) NQG + EL (pre)}: \textit{NQG + Entity Linking} with the pre-trained entity linking feature obtained from the joint training of word and Wikipedia entity using \cite{yamada2016joint}.\\
\textbf{(5) NQG + FGET}: Extension of \textit{NQG} with the fine grained entity type (FGET) feature (100 dimension) discussed in section \ref{fgec}.\\
\textbf{(6) NQG + FGET (pre)}: \textit{NQG + FGET} with the pre-trained FGET features as discussed in Section \ref{fgec}.\\
\textbf{(7) NQG + EL (pre) + FGET (pre)}: Combination of \textit{NQG}, \textit{Entity Linking} and \textit{FGET} with pre-trained entity linking and FGET features.\\
\indent In order to compare our models with the existing coarse-grained entity features (NER) being used in literature \cite{zhou2017neural,harrison2018neural}, we also report the following experiments.\\
\textbf{(1) NQG + NER}: \textit{NQG} with the coarse-grained named entity recognition\footnote{We use the Stanford NER \cite{finkel2005incorporating} to tag the entity} feature.\\
\textbf{(2) NQG + NER + FGET}: \textit{NQG}, \textit{NER} and \textit{FGET} with NER (100 dimension) and FGET features\\
\textbf{(3) NQG + NER + FGET (pre)}: \textit{NQG}, \textit{NER} and \textit{FGET} with NER (100 dimension) and pre-trained FGET features \\
We report the results on the test set of SQuAD and MS MARCO in Table \ref{table:test-results}. The results on the development set of both datasets are shown in Table \ref{table:dev-results}. 
\subsection{Discussion and Analysis}
Table  \ref{table:test-results} clearly demonstrates that the proposed fine-grained word-knowledge features improve the performance of the models over the baseline, and the coarse-grained entity (NER) features do not seem to be as useful as the entity linking features for both datasets. We analyzed the effect of each word-knowledge feature on both datasets. Our findings are as follows:  
\paragraph{Entity Linking:} On both datasets, the pretrained entity linking features were more effective compared to randomly initialized features followed by fine-tuning while training. We believe this is due to the word and corresponding entity being jointly trained and projected into the same vector space. We observe that entity linking features on SQuAD are less effective than MS MARCO. 
\paragraph{FGET:}
Similar to the linker-based features, the pretrained FGET features trained on the FIGER dataset \cite{ling2012fine} are more effective than the randomly initialized vectors. The FGET feature is more effective at improving the QG model on SQuAD. We believe this is likely because both the SQuAD and FIGER datasets were derived from Wikipedia. In contrast, MS MARCO was derived from Bing\footnote{\url{https://www.bing.com/}} user queries and web passages, which are entirely different in nature. It should also be noted that the FGET features were derived using entities detected using the entity linker. In order to evaluate the effect of using the linker as an entity detector, we also performed an experiment in which we used entities detected using the NER. We found that the models that use the entities detected with the linker have higher performance in terms of each evaluation metric on both datasets.
\section{Conclusion and Future Work}
We proposed that features based on general word-knowledge can improve the performance of question generation. Our results on SQuAD and MS MARCO show that entity-based world knowledge is effective at improving question generation according to automated metrics. In order to fully explore the performance gains of these features, human evaluation is required, and we leave this for future work. We would also like to explore other sources of world knowledge beyond entity-based information. In particular, we believe that information based on the relationships between the entities present in the passage would also be useful. 
\bibliography{acl2019}

\begin{thebibliography}{48}
\expandafter\ifx\csname natexlab\endcsname\relax\def\natexlab#1{#1}\fi

\bibitem[{Agarwal et~al.(2025)Agarwal, Akhtar, and Yadav}]{agarwal2025overview}
Siddhant Agarwal, Md~Shad Akhtar, and Shweta Yadav. 2025.
\newblock Overview of the peranssumm 2025 shared task on perspective-aware
  healthcare answer summarization.
\newblock In \emph{Proceedings of the Second Workshop on Patient-Oriented
  Language Processing (CL4Health)}, pages 445--455.

\bibitem[{Bahdanau et~al.(2014)Bahdanau, Cho, and Bengio}]{bahdanau2014neural}
Dzmitry Bahdanau, Kyunghyun Cho, and Yoshua Bengio. 2014.
\newblock {Neural Machine Translation by Jointly Learning to Align and
  Translate}.
\newblock \emph{arXiv preprint arXiv:1409.0473}.

\bibitem[{Bahri et~al.(2024)Bahri, Oueslati, and
  Kharchoufi}]{bahri2024consumer}
Afef Bahri, Wided Oueslati, and Abdelkarim Kharchoufi. 2024.
\newblock Consumer health question summarization using transformers and data
  augmentation.
\newblock In \emph{International Conference on Science, Engineering Management
  and Information Technology}, pages 244--255. Springer.

\bibitem[{Banerjee and Lavie(2005)}]{meteor}
Satanjeev Banerjee and Alon Lavie. 2005.
\newblock \href {http://aclweb.org/anthology/W05-0909} {Meteor: An automatic
  metric for mt evaluation with improved correlation with human judgments}.
\newblock In \emph{Proceedings of the ACL Workshop on Intrinsic and Extrinsic
  Evaluation Measures for Machine Translation and/or Summarization}, pages
  65--72. Association for Computational Linguistics.

\bibitem[{Bhattacharya et~al.(2022)Bhattacharya, Chaturvedi, and
  Yadav}]{bhattacharya2022lchqa}
Abari Bhattacharya, Rochana Chaturvedi, and Shweta Yadav. 2022.
\newblock Lchqa-summ: Multi-perspective summarization of publicly sourced
  consumer health answers.
\newblock In \emph{Proceedings of the First Workshop on Natural Language
  Generation in Healthcare}, pages 23--26.

\bibitem[{Chali and Hasan(2015)}]{chali2015towards}
Yllias Chali and Sadid~A Hasan. 2015.
\newblock Towards topic-to-question generation.
\newblock \emph{Computational Linguistics}, 41(1):1--20.

\bibitem[{Chaturvedi et~al.(2024)Chaturvedi, Bhattacharya, and
  Yadav}]{chaturvedi2024aspect}
Rochana Chaturvedi, Abari Bhattacharya, and Shweta Yadav. 2024.
\newblock Aspect-oriented consumer health answer summarization.
\newblock \emph{arXiv preprint arXiv:2405.06295}.

\bibitem[{Cheng and Roth(2013)}]{cheng2013relational}
Xiao Cheng and Dan Roth. 2013.
\newblock Relational inference for wikification.
\newblock In \emph{Proceedings of the 2013 Conference on Empirical Methods in
  Natural Language Processing}, pages 1787--1796.

\bibitem[{Du et~al.(2017)Du, Shao, and Cardie}]{learning-to-ask}
Xinya Du, Junru Shao, and Claire Cardie. 2017.
\newblock \href {https://doi.org/10.18653/v1/P17-1123} {Learning to ask: Neural
  question generation for reading comprehension}.
\newblock In \emph{Proceedings of the 55th Annual Meeting of the Association
  for Computational Linguistics (Volume 1: Long Papers)}, pages 1342--1352.
  Association for Computational Linguistics.

\bibitem[{Finkel et~al.(2005)Finkel, Grenager, and
  Manning}]{finkel2005incorporating}
Jenny~Rose Finkel, Trond Grenager, and Christopher Manning. 2005.
\newblock Incorporating non-local information into information extraction
  systems by gibbs sampling.
\newblock In \emph{Proceedings of the 43rd annual meeting on association for
  computational linguistics}, pages 363--370. Association for Computational
  Linguistics.

\bibitem[{Gao et~al.(2018)Gao, Wang, Bing, King, and Lyu}]{gao2018difficulty}
Yifan Gao, Jianan Wang, Lidong Bing, Irwin King, and Michael~R Lyu. 2018.
\newblock Difficulty controllable question generation for reading
  comprehension.
\newblock \emph{arXiv preprint arXiv:1807.03586}.

\bibitem[{Gulcehre et~al.(2016)Gulcehre, Ahn, Nallapati, Zhou, and
  Bengio}]{pointing-unknown-gulcehre-2016}
Caglar Gulcehre, Sungjin Ahn, Ramesh Nallapati, Bowen Zhou, and Yoshua Bengio.
  2016.
\newblock \href {https://doi.org/10.18653/v1/P16-1014} {Pointing the unknown
  words}.
\newblock In \emph{Proceedings of the 54th Annual Meeting of the Association
  for Computational Linguistics (Volume 1: Long Papers)}, pages 140--149.
  Association for Computational Linguistics.

\bibitem[{Harrison and Walker(2018)}]{harrison2018neural}
Vrindavan Harrison and Marilyn Walker. 2018.
\newblock Neural generation of diverse questions using answer focus, contextual
  and linguistic features.
\newblock \emph{arXiv preprint arXiv:1809.02637}.

\bibitem[{Heilman and Smith(2010)}]{heilman2010good}
Michael Heilman and Noah~A Smith. 2010.
\newblock {Good Question! Statistical Ranking for Question Generation}.
\newblock In \emph{Human Language Technologies: The 2010 Annual Conference of
  the North American Chapter of the Association for Computational Linguistics},
  pages 609--617. Association for Computational Linguistics.

\bibitem[{Hochreiter and Schmidhuber(1997)}]{hochreiter1997long}
Sepp Hochreiter and J{\"u}rgen Schmidhuber. 1997.
\newblock {Long Short-Term Memory}.
\newblock \emph{Neural computation}, 9(8):1735--1780.

\bibitem[{Kim et~al.(2018)Kim, Lee, Shin, and Jung}]{kim2018improving}
Yanghoon Kim, Hwanhee Lee, Joongbo Shin, and Kyomin Jung. 2018.
\newblock Improving neural question generation using answer separation.
\newblock \emph{arXiv preprint arXiv:1809.02393}.

\bibitem[{Kingma and Ba(2014)}]{kingma2014adam}
Diederik~P Kingma and Jimmy Ba. 2014.
\newblock {Adam: A Method for Stochastic Optimization}.
\newblock \emph{arXiv preprint arXiv:1412.6980}.

\bibitem[{Kumar et~al.(2016)Kumar, Irsoy, Ondruska, Iyyer, Bradbury, Gulrajani,
  Zhong, Paulus, and Socher}]{kumar2016ask}
Ankit Kumar, Ozan Irsoy, Peter Ondruska, Mohit Iyyer, James Bradbury, Ishaan
  Gulrajani, Victor Zhong, Romain Paulus, and Richard Socher. 2016.
\newblock {Ask Me Anything: Dynamic Memory Networks for Natural Language
  Processing}.
\newblock In \emph{International Conference on Machine Learning}, pages
  1378--1387.

\bibitem[{Labutov et~al.(2015)Labutov, Basu, and Vanderwende}]{labutov2015deep}
Igor Labutov, Sumit Basu, and Lucy Vanderwende. 2015.
\newblock Deep questions without deep understanding.
\newblock In \emph{Proceedings of the 53rd Annual Meeting of the Association
  for Computational Linguistics and the 7th International Joint Conference on
  Natural Language Processing (Volume 1: Long Papers)}, volume~1, pages
  889--898.

\bibitem[{Lin(2004)}]{lin2004rouge}
Chin-Yew Lin. 2004.
\newblock {ROUGE: A Package for Automatic Evaluation of Summaries}.
\newblock In \emph{Text Summarization Branches Out: Proceedings of the ACL-04
  Workshop}, volume~8. Barcelona, Spain.

\bibitem[{Lindberg et~al.(2013)Lindberg, Popowich, Nesbit, and
  Winne}]{lindberg2013generating}
David Lindberg, Fred Popowich, John Nesbit, and Phil Winne. 2013.
\newblock Generating natural language questions to support learning on-line.
\newblock In \emph{Proceedings of the 14th European Workshop on Natural
  Language Generation}, pages 105--114.

\bibitem[{Ling and Weld(2012)}]{ling2012fine}
Xiao Ling and Daniel~S Weld. 2012.
\newblock Fine-grained entity recognition.
\newblock In \emph{AAAI}, volume~12, pages 94--100.

\bibitem[{Luong et~al.(2015)Luong, Pham, and Manning}]{luong-attention}
Thang Luong, Hieu Pham, and Christopher~D. Manning. 2015.
\newblock \href {https://doi.org/10.18653/v1/D15-1166} {Effective approaches to
  attention-based neural machine translation}.
\newblock In \emph{Proceedings of the 2015 Conference on Empirical Methods in
  Natural Language Processing}, pages 1412--1421. Association for Computational
  Linguistics.

\bibitem[{Naik et~al.(2024)Naik, Chandakacherla, Yadav, and
  Akhtar}]{naik2024no}
Gauri Naik, Sharad Chandakacherla, Shweta Yadav, and Md~Shad Akhtar. 2024.
\newblock No perspective, no perception!! perspective-aware healthcare answer
  summarization.
\newblock In \emph{Findings of the Association for Computational Linguistics
  ACL 2024}, pages 15919--15932.

\bibitem[{Nguyen et~al.(2016)Nguyen, Rosenberg, Song, Gao, Tiwary, Majumder,
  and Deng}]{ms-marco}
Tri Nguyen, Mir Rosenberg, Xia Song, Jianfeng Gao, Saurabh Tiwary, Rangan
  Majumder, and Li~Deng. 2016.
\newblock \href
  {https://www.microsoft.com/en-us/research/publication/ms-marco-human-generated-machine-reading-comprehension-dataset/}
  {Ms marco: A human generated machine reading comprehension dataset}.

\bibitem[{Papineni et~al.(2002)Papineni, Roukos, Ward, and
  Zhu}]{papineni2002bleu}
Kishore Papineni, Salim Roukos, Todd Ward, and Wei-Jing Zhu. 2002.
\newblock {BLEU: A Method for Automatic Evaluation of Machine Translation}.
\newblock In \emph{Proceedings of the 40th Annual Meeting on Association for
  Computational Linguistics}, pages 311--318. Association for Computational
  Linguistics.

\bibitem[{Rajpurkar et~al.(2016)Rajpurkar, Zhang, Lopyrev, and Liang}]{squad}
Pranav Rajpurkar, Jian Zhang, Konstantin Lopyrev, and Percy Liang. 2016.
\newblock \href {https://doi.org/10.18653/v1/D16-1264} {{SQuAD: 100,000+
  Questions for Machine Comprehension of Text}}.
\newblock In \emph{Proceedings of the 2016 Conference on Empirical Methods in
  Natural Language Processing}, pages 2383--2392. Association for Computational
  Linguistics.

\bibitem[{Rush et~al.(2015)Rush, Chopra, and Weston}]{rush2015neural}
Alexander~M Rush, Sumit Chopra, and Jason Weston. 2015.
\newblock A neural attention model for abstractive sentence summarization.
\newblock \emph{arXiv preprint arXiv:1509.00685}.

\bibitem[{See et~al.(2017)See, Liu, and Manning}]{go-to-the-point-see-2017}
Abigail See, Peter~J. Liu, and Christopher~D. Manning. 2017.
\newblock \href {https://doi.org/10.18653/v1/P17-1099} {Get to the point:
  Summarization with pointer-generator networks}.
\newblock In \emph{Proceedings of the 55th Annual Meeting of the Association
  for Computational Linguistics (Volume 1: Long Papers)}, pages 1073--1083.
  Association for Computational Linguistics.

\bibitem[{Sharma et~al.(2017)Sharma, El~Asri, Schulz, and
  Zumer}]{sharma2017nlgeval}
Shikhar Sharma, Layla El~Asri, Hannes Schulz, and Jeremie Zumer. 2017.
\newblock \href {http://arxiv.org/abs/1706.09799} {Relevance of unsupervised
  metrics in task-oriented dialogue for evaluating natural language
  generation}.
\newblock \emph{CoRR}, abs/1706.09799.

\bibitem[{Song et~al.(2018)Song, Wang, Hamza, Zhang, and
  Gildea}]{context-multi-perspective-qg}
Linfeng Song, Zhiguo Wang, Wael Hamza, Yue Zhang, and Daniel Gildea. 2018.
\newblock \href {https://doi.org/10.18653/v1/N18-2090} {Leveraging context
  information for natural question generation}.
\newblock In \emph{Proceedings of the 2018 Conference of the North American
  Chapter of the Association for Computational Linguistics: Human Language
  Technologies, Volume 2 (Short Papers)}, pages 569--574. Association for
  Computational Linguistics.

\bibitem[{Srivastava et~al.(2015)Srivastava, Mansimov, and
  Salakhudinov}]{srivastava2015unsupervised}
Nitish Srivastava, Elman Mansimov, and Ruslan Salakhudinov. 2015.
\newblock Unsupervised learning of video representations using lstms.
\newblock In \emph{International conference on machine learning}, pages
  843--852.

\bibitem[{Sun et~al.(2018)Sun, Liu, Lyu, He, Ma, and Wang}]{ans-focused-qg}
Xingwu Sun, Jing Liu, Yajuan Lyu, Wei He, Yanjun Ma, and Shi Wang. 2018.
\newblock \href {http://aclweb.org/anthology/D18-1427} {Answer-focused and
  position-aware neural question generation}.
\newblock In \emph{Proceedings of the 2018 Conference on Empirical Methods in
  Natural Language Processing}, pages 3930--3939. Association for Computational
  Linguistics.

\bibitem[{Sutskever et~al.(2014)Sutskever, Vinyals, and
  Le}]{sutskever2014sequence}
Ilya Sutskever, Oriol Vinyals, and Quoc~V Le. 2014.
\newblock {Sequence to Sequence Learning with Neural Networks}.
\newblock In \emph{Advances in Neural Information Processing Systems}, pages
  3104--3112.

\bibitem[{Trischler et~al.(2016)Trischler, Wang, Yuan, Harris, Sordoni,
  Bachman, and Suleman}]{trischler2016newsqa}
Adam Trischler, Tong Wang, Xingdi Yuan, Justin Harris, Alessandro Sordoni,
  Philip Bachman, and Kaheer Suleman. 2016.
\newblock {NewsQA: A Machine Comprehension Dataset}.
\newblock \emph{arXiv preprint arXiv:1611.09830}.

\bibitem[{Wang et~al.(2016)Wang, Mi, Hamza, and Florian}]{wang2016multi}
Zhiguo Wang, Haitao Mi, Wael Hamza, and Radu Florian. 2016.
\newblock Multi-perspective context matching for machine comprehension.
\newblock \emph{arXiv preprint arXiv:1612.04211}.

\bibitem[{Xu et~al.(2015)Xu, Ba, Kiros, Cho, Courville, Salakhudinov, Zemel,
  and Bengio}]{xu2015show}
Kelvin Xu, Jimmy Ba, Ryan Kiros, Kyunghyun Cho, Aaron Courville, Ruslan
  Salakhudinov, Rich Zemel, and Yoshua Bengio. 2015.
\newblock Show, attend and tell: Neural image caption generation with visual
  attention.
\newblock In \emph{International conference on machine learning}, pages
  2048--2057.

\bibitem[{Xu and Barbosa(2018)}]{fgec}
Peng Xu and Denilson Barbosa. 2018.
\newblock \href {http://aclweb.org/anthology/N18-1002} {Neural fine-grained
  entity type classification with hierarchy-aware loss}.
\newblock In \emph{Proceedings of the 2018 Conference of the North American
  Chapter of the Association for Computational Linguistics: Human Language
  Technologies, Volume 1 (Long Papers)}, pages 16--25. Association for
  Computational Linguistics.

\bibitem[{Yadav and Caragea(2022)}]{yadav2022towards}
Shweta Yadav and Cornelia Caragea. 2022.
\newblock Towards summarizing healthcare questions in low-resource setting.
\newblock In \emph{Proceedings of the 29th International Conference on
  Computational Linguistics}, pages 2892--2905.

\bibitem[{Yadav et~al.(2023)Yadav, Cobeli, and Caragea}]{yadav2023towards}
Shweta Yadav, Stefan Cobeli, and Cornelia Caragea. 2023.
\newblock Towards understanding consumer healthcare questions on the web with
  semantically enhanced contrastive learning.
\newblock In \emph{Proceedings of the ACM Web Conference 2023}, pages
  1773--1783.

\bibitem[{Yadav et~al.(2021{\natexlab{a}})Yadav, Gupta, Abacha, and
  Demner-Fushman}]{yadav2021reinforcement}
Shweta Yadav, Deepak Gupta, Asma~Ben Abacha, and Dina Demner-Fushman.
  2021{\natexlab{a}}.
\newblock Reinforcement learning for abstractive question summarization with
  question-aware semantic rewards.
\newblock In \emph{Proceedings of the 59th Annual Meeting of the Association
  for Computational Linguistics and the 11th International Joint Conference on
  Natural Language Processing (Volume 2: Short Papers)}, pages 249--255.

\bibitem[{Yadav et~al.(2022{\natexlab{a}})Yadav, Gupta, Abacha, and
  Demner-Fushman}]{yadav2022question}
Shweta Yadav, Deepak Gupta, Asma~Ben Abacha, and Dina Demner-Fushman.
  2022{\natexlab{a}}.
\newblock Question-aware transformer models for consumer health question
  summarization.
\newblock \emph{Journal of Biomedical Informatics}, 128:104040.

\bibitem[{Yadav et~al.(2022{\natexlab{b}})Yadav, Gupta, and
  Demner-Fushman}]{yadav2022chq}
Shweta Yadav, Deepak Gupta, and Dina Demner-Fushman. 2022{\natexlab{b}}.
\newblock Chq-summ: A dataset for consumer healthcare question summarization.
\newblock \emph{arXiv preprint arXiv:2206.06581}.

\bibitem[{Yadav et~al.(2021{\natexlab{b}})Yadav, Sarrouti, and
  Gupta}]{yadav2021nlm}
Shweta Yadav, Mourad Sarrouti, and Deepak Gupta. 2021{\natexlab{b}}.
\newblock Nlm at mediqa 2021: Transfer learning-based approaches for consumer
  question and multi-answer summarization.
\newblock In \emph{proceedings of the 20th workshop on biomedical language
  processing}, pages 291--301.

\bibitem[{Yamada et~al.(2016)Yamada, Shindo, Takeda, and
  Takefuji}]{yamada2016joint}
Ikuya Yamada, Hiroyuki Shindo, Hideaki Takeda, and Yoshiyasu Takefuji. 2016.
\newblock Joint learning of the embedding of words and entities for named
  entity disambiguation.
\newblock \emph{arXiv preprint arXiv:1601.01343}.

\bibitem[{Zhao et~al.(2024)Zhao, Deng, Yadav, and Yu}]{zhao2024heterogeneous}
Wenting Zhao, Zhongfen Deng, Shweta Yadav, and Philip~S Yu. 2024.
\newblock Heterogeneous knowledge grounding for medical question answering with
  retrieval augmented large language model.
\newblock In \emph{Companion Proceedings of the ACM Web Conference 2024}, pages
  1590--1594.

\bibitem[{Zhao et~al.(2018)Zhao, Ni, Ding, and Ke}]{para-level-max-out-qg}
Yao Zhao, Xiaochuan Ni, Yuanyuan Ding, and Qifa Ke. 2018.
\newblock \href {http://aclweb.org/anthology/D18-1424} {Paragraph-level neural
  question generation with maxout pointer and gated self-attention networks}.
\newblock In \emph{Proceedings of the 2018 Conference on Empirical Methods in
  Natural Language Processing}, pages 3901--3910. Association for Computational
  Linguistics.

\bibitem[{Zhou et~al.(2017)Zhou, Yang, Wei, Tan, Bao, and
  Zhou}]{zhou2017neural}
Qingyu Zhou, Nan Yang, Furu Wei, Chuanqi Tan, Hangbo Bao, and Ming Zhou. 2017.
\newblock Neural question generation from text: A preliminary study.
\newblock In \emph{National CCF Conference on Natural Language Processing and
  Chinese Computing}, pages 662--671. Springer.

\end{thebibliography}
\bibliographystyle{acl_natbib}
\end{document}